# Conditional Diffusion Model for Longitudinal Medical Image Generation

Duy-Phuong Dao
Department of Artificial Intelligence Convergence
Chonnam National University
Gwangju, South Korea
phuongdd.1997@gmail.com

Hyung-Jeong Yang*
Department of Artificial Intelligence Convergence
Chonnam National University
Gwangju, South Korea
hjyang@jnu.ac.kr

Jahae Kim
Department of Nuclear Medicine & Department of Artificial Intelligence Convergence, Chonnam National University
Gwangju, South Korea
jhbt0607@hanmail.net

*Abstract*—Alzheimer's disease progresses slowly and involves complex interaction between various biological factors. Longitudinal medical imaging data can capture this progression over time. However, longitudinal data frequently encounter issues such as missing data due to patient dropouts, irregular follow-up intervals, and varying lengths of observation periods. To address these issues, we designed a diffusion-based model for 3D longitudinal medical imaging generation using single magnetic resonance imaging (MRI). This involves the injection of a conditioning MRI and time-visit encoding to the model, enabling control in change between source and target images. The experimental results indicate that the proposed method generates higher-quality images compared to other competing methods.

## I. INTRODUCTION

In a recent decade, the advent of generative artificial intelligent (AI) models has opened up new paths in various fields. Generative AI, notably through models like generative adversarial networks (GANs) [1-3], and variational autoencoders (VAEs) [4, 5] have demonstrated remarkable capabilities in generating synthetic medical imaging data. However, it is difficult and unstable to train GANs due to involving a delicate balance between the generation and discriminator. And the generator of GAN also produces a limited variety of outputs despite diverse training data. As for VAE, this mechanism tends to produce images with blurry details due to its reliance Gaussian assumptions in the latent space.

Fortunately, diffusion methods [6-8] recently offer a promising alterative, addressing these limitations with stable training processes, high-resolution outputs, and enhanced robustness. While diffusion models have shown promise in generating medical images, including image registration [9], segmentation [10], or super-resolution [11], there have not been many studies focusing on longitudinal medical image generation. This is particularly important as longitudinal data plays a vital role in predicting disease progression and treatment planning. However, due to the cost and potential harm, it is hard and takes long time to acquire enough data for prediction. Therefore, we propose a conditional diffusion model for generating longitudinal magnetic resonance imaging (MRI) scans using only a baseline MRI scan. We conditions the diffusion model by using a source image and time interval.

The organization of the subsequent sections of this paper is as follows: Section II reviews prior research on medical image generation. Section III describes the proposed model. Section IV discusses the dataset and experimental results, including a comparison with existing methods. Lastly, Section V summarizes our findings and conclusions.

## II. RELATED WORK

VAEs and GANs were the most widely used image generation methods before the emergence of diffusion models. Nouveau VAE [12], a hierarchical VAE aimed at generating realistic images, was introduced by Vahdat and Kautz. VQ-VAE [13] employed vector quantization to capture intricate details through discrete latent variables. G. Kwon et al. [14] pioneered the concept of generating 3D brain MRIs using an auto-encoding GAN. L. Sun et al. [15] utilized a hierarchical amortized GAN to generate high-resolution 3D medical images. However, these aforementioned methods face challenges related to training instability.

Recently, 3D-DDPM [16] was proposed to generate 3D synthetic brain MRIs by using diffusion. L. Pinaya et al. [17] incorporated multiple variables as conditions and employed latent diffusion to create 3D realistic brain MRIs with specific covariate values. However, these methods did not specifically target the generation of longitudinal medical images.

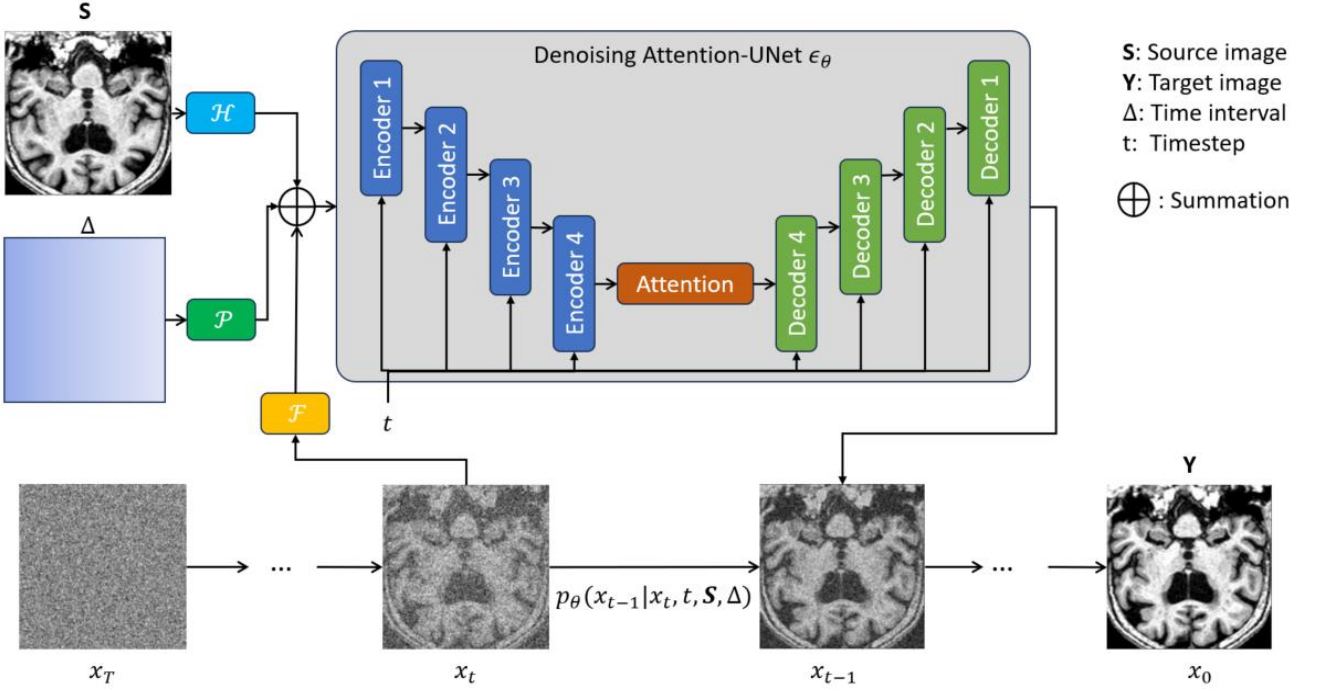

Figure 1. The overall architecture of the proposed method.

## III. PROPOSED METHOD

Fig. 1 illustrates the overall architecture of the proposed method. Given the source image S and the time interval Δ, we design a diffusion model to generate the target image Y that represents the evolved state of S after time Δ. We propose a conditional diffusion module to capture the intermediate transitions between the target and source images.

### A. Diffusion background

We briefly present an overview of the operational principles behind the diffusion method from [1]. The diffusion method is composed of two primary processes: forward diffusion process q and reverse diffusion process $p_\theta$. Given an image $x_0$ and a fixed value $\beta_t$ that determines the pattern of added noise, we can sample a noisy image $x_t$ at any timestamp t during the forward process $q$. This process is outlined as follows:

$$q(x_t|x_0) = \mathcal{N}(x_t; \sqrt{\bar{\alpha}_t}x_0, (1-\bar{\alpha}_t)\mathbf{I}), \quad 0<t<T \quad (1)$$

which can be expressed differently as:

$$x_t = \sqrt{\bar{\alpha}_t}x_0 + \sqrt{(1-\bar{\alpha}_t)}\epsilon, \quad \epsilon \sim \mathcal{N}(0,\mathbf{I}) \quad (2)$$

where $\alpha_t = 1 - \beta_t$, $\bar{\alpha}_t = \prod_{k=1}^{t}\alpha_k$, T is the number of diffusion steps, $\epsilon$ is a random noise sample from Gaussian distribution with a mean of zero and an identity covariance matrix **I**.

After T diffusion steps, the image $x_0$ is transformed into $x_T$ that closely resembles a Gaussian noise with a mean of zeros and a standard deviation of one. From the noisy image $x_T$, the reverse process $p_\theta$ from timestamp $t$ to $t-1$ is formulated as follows:

$$p_\theta(x_t) = \mathcal{N}(x_T; \mathbf{0},\mathbf{I}) \quad (3)$$

$$p_\theta(x_{t-1}|x_t) = \mathcal{N}(x_{t-1}; \mu_\theta(x_t,t), \Sigma_\theta(x_t,t)\mathbf{I}) \quad (4)$$

where $\mu_\theta(x_t,t)$ is trainable mean and $\Sigma_\theta(x_t,t)$ is set to constant at each timestamp t. Following [1], $\mu_\theta(x_t,t)$ and $\Sigma_\theta(x_t,t)$ are formulated as follows:

$$\mu_\theta(x_t,t) = \frac{1}{\sqrt{\alpha_t}}\left(x_t - \frac{\beta_t}{\sqrt{1-\bar{\alpha}_t}}\epsilon_\theta(x_t,t)\right) \quad (5)$$

$$\Sigma_\theta(x_t,t) = \frac{1-\bar{\alpha}_{t-1}}{1-\bar{\alpha}_t}\beta_t \quad (6)$$

Finally, the variational lower bound is performed to optimize the diffusion models, described as follows:

$$\mathcal{L} = \sum_{1<t<T} D_{KL}(q(x_{t-1}|x_0,x_t) \| p_\theta(x_{t-1}|x_t)) \quad (7)$$

where $D_{KL}$ is Kullback-Leibler divergence loss and $q(x_{t-1}|x_t,x_0)$ is defined using Bayes theorem as follows:

$$q(x_{t-1}|x_t,x_0) = \mathcal{N}(x_{t-1}; \tilde{\mu}(x_t,x_0), \frac{1-\bar{\alpha}_{t-1}}{1-\bar{\alpha}_t}\beta_t\mathbf{I}) \quad (8)$$

$$\tilde{\mu}(x_t,x_0) = \frac{\sqrt{\bar{\alpha}_{t-1}}\beta_t}{1-\bar{\alpha}_t}x_0 + \frac{\sqrt{\alpha_t}(1-\bar{\alpha}_{t-1})}{1-\bar{\alpha}_t}x_t \quad (9)$$

Therefore, $\mathcal{L}$ in equation () is parameterized as follows:

$$\mathcal{L} = \left\|\epsilon - \epsilon_\theta(\sqrt{\bar{\alpha}_t}x_0 + \sqrt{(1-\bar{\alpha}_t)}\epsilon, t)\right\|^2 \quad (10)$$

### B. Conditional Diffusion Module

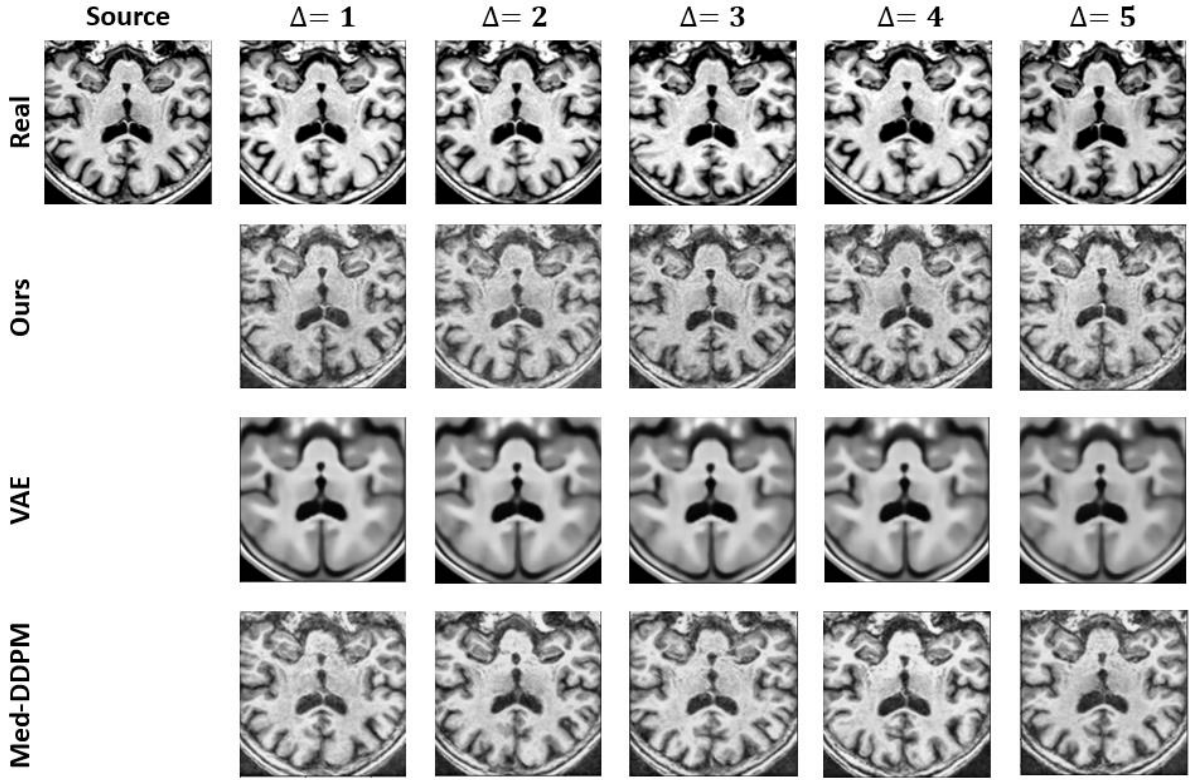

Figure 2. Qualitative comparison of overall quality in the longitudinal MRI generation.

We refine the reverse diffusion module by injecting source image S and the time interval $\Delta$ into $\epsilon_\theta$, along with the current estimate $x_t$ and timestep t. Inspired by the effective of residual connections [18], we aggregate these conditions through addition operation. The aggregated information is then fed into attention-based UNet [19] model to estimate the added noise. Lastly, we repeat the reverse process T times to guide the model in generating the target image Y. The formulation of $\epsilon_\theta$ at timestep t can be expressed as follows:

$$\epsilon_\theta(x_t, t, S, \Delta) = \mathcal{D}(\mathcal{E}(\mathcal{F}(x_t) + \mathcal{H}(S) + \mathcal{P}(\Delta), t), t) \quad (11)$$

where $\mathcal{H}$ and $\mathcal{F}$ are convolutional networks, $\mathcal{E}$ and $\mathcal{D}$ serve as the encoder and decoder of the attention-based UNet model, respectively. It's worth noting that conditional module $\epsilon_\theta(x_t, t, S, \Delta)$ in equation (11) is integrated into equation (5), substituting for the unconditional module $\epsilon_\theta(x_t, t)$.

## IV. EXPERIMENTAL SETTINGS AND RESULTS

### A. Dataset

In this study, we leveraged The Alzheimer's Disease Neuroimaging Initiative (ADNI)[1] dataset for evaluating our proposed method. This dataset includes data collected from participants over multiple time points, typically spanning several years. We considered 1239 participants who had undergone MRI scans at least twice. Out of these, we used 1,125 participants for training and the remaining participants for evaluation. The interval time between two consecutive visits is one year. We selected the MRI scan from the first visit as the source image and the scans from the follow-up visits as the target images. The input size is 128×160×128. We also normalized the image to range from -1 to 1.

TABLE I. QUANTITATIVE COMPARISON

| Method | FID↓ | SSIM↑ |
|---|---|---|
| VAE | 201.32 | 0.2662 |
| Med-DDPM [22] | 39.28 | 0.2314 |
| Ours | **33.75** | **0.2774** |

### B. Training Settings

We conducted comprehensive experiments with T=1000. The $\beta_t$ were set to increase linearly from $10^{-4}$ at t=1 to 0.02 at t=T. The Adam optimizer [20] was utilized to update the model's trainable parameters with a fixed learning rate of 0.001. A total of 300,000 training iterations were conducted, employing a batch size of 1 for each update.

### C. Experimental Results

We evaluated the performance of generating the longitudinal MRI data by estimating Fréchet inception distance (FID) [21] and structural similarity (SSIM) metrics. The Table I shows that the proposed method has better results compared with other methods in terms of the all evaluation metrics. We also visualized several generated samples of MRI scans in five followup visits using the MRI scan of the current visit. As shown in Figure 2, the quality of the generated images by the proposed method is higher than that of the competing methods.

## V. CONCLUSION

In this work, we proposed a conditional diffusion model for generating the longitudinal medical image data using only a current image. Our model combined the current image and time interval and then injects them into the reverse diffusion process as conditional term. The quantitative and qualitative results show that the proposed method is better than the competing methods. However, we have not yet leveraged the temporal information between observed images to generate subsequent images. In future work, we will focus on this temporal information to improve the continuity between the source and target images.

## *Acknowledgment*

This work was supported by Institute of Information & communications Technology Planning & Evaluation (IITP) under the Artificial Intelligence Convergence Innovation Human Resources Development (IITP-2023-RS-2023-00256629) grant funded by the Korea government(MSIT).

This work was supported by the National Research Foundation of Korea(NRF) grant funded by the Korea government(MSIT). (RS-2023-00208397).